\documentstyle[aclap]{article}
\title{\vspace{-0.5in}Automatic Extraction of
Tagset Mappings from Parallel-Annotated Corpora}
\author{John Hughes, Clive Souter \and Eric Atwell \\
Centre for the Computer Analysis of Language And Speech \\
School of Computer Studies, Leeds University, Leeds LS2 9JT,  UK \\
{\tt john@scs.leeds.ac.uk, cs@scs.leeds.ac.uk, eric@scs.leeds.ac.uk}}

\begin{document}

\maketitle
\vspace{-0.5in}

\begin{abstract}

Several research projects around the world are building grammatically
analysed corpora; that is, collections of text annotated with part-of-speech
wordtags and syntax trees. However, projects have used quite different
wordtagging and parsing schemes.
Developers of corpora adhere to a variety of competing
models or theories of grammar and parsing, with the effect of restricting the
accessibility of their respective corpora, and the potential for collation
into a single fully parsed corpus.
In view of this heterogeneity, we have begun to investigate
and develop methods
of automatically mapping between the annotation schemes of the most widely
known corpora, thus assessing their differences and improving their
reusability.
Annotating a single corpus with the different schemes allows
for comparisons and will provide a rich test-bed for automatic parsers.
Collation of all the included corpora into a single large annotated corpus
will provide a more detailed language model to be developed for tasks
such as speech and handwriting recognition.
This paper focuses on methods of developing mappings between tagsets and,
in particular, the method of automatic extraction of mappings from corpora
tagged with more than one annotation scheme.

\end{abstract}

\vspace{0.02in}
\section {Introduction}

Many, diverse tagged and parsed corpora have been developed.  Amongst the
applications of annotated corpora are as training sets for the extraction
of models used in speech and handwriting recognition.  Such training sets
need to be as large as possible and there is anecdotal
evidence that even the largest on its own is too small for
a general statistical model of higher-level
syntactic structure.  As annotating corpora using
hand-crafted markup or some semi-automated process followed by correction
by linguistic experts is slow and expensive \cite{Barkema93,LeechGarside91}
it would be preferable if some other method of building a large
annotated corpus could be found.
Existing corpora were not designed to a specific
framework of annotations so corpora can not easily be collated into a
single large training set.  The AMALGAM (automatic mapping among
lexico-grammatical annotation models) project was set up to research ways
of mapping between annotation schemes in order to increase the size of corpus
tagged with the schemes included in the project
\cite{AtwellHughesSouter94a,AtwellHughesSouter94b}.

We are developing
a multi-tagged corpus and a multi-treebank, a single text-set annotated
with all the tagging and parsing schemes we include in the mappings.
The text-set is the Spoken
English Corpus (SEC); which is already annotated with two syntax schemes.
However, the main
deliverable to the computational linguistics research community is not the
SEC-based multi-treebank, but its associated suite of mappings - this can
be used to combine currently-incompatible syntactic training sets into a large
unified corpus.  Our development of the mapping algorithms aims to
distinguish notational from substantive differences in the annotation schemes,
and we will be able to evaluate tagging schemes in terms of how well they fit
standard statistical language models such as n-pos (Markov) models.

Although the above description assumes mapping between tagsets from
monolingual corpora we believe the issues extend to multilingual tagsets.
The tagsets of two languages usually differ in the features they cover.
For example French may have tags to discriminate gender whereas English does
not.  However, tagsets of English do not necessarily mutually cover all
features.  For instance, the British component of the International Corpus
of English \cite{Greenbaum93} has a tagging scheme
that accounts for transitivity of verbs whereas the Lancaster/Oslo Bergen
corpus \cite{JohanssonAtwellGarsideLeech86}
does not (nor do the EAGLES proposals - see below).  We
believe that our methods are scalable to mappings between
multilingual tagsets.

\section{Related Research}

Corpus-trained statistical language learning techniques have been
successfully applied to a range of problems in computational linguistics,
including part-of-speech wordtagging
\cite{LeechGarsideAtwell83,Atwell83,Atwell87a},
word sense disambiguation and tagging \cite{Demetriou93,Galeetal92},
learning word classes \cite{Atwell87b,AtwellDrakos87,Hughes93,HughesAtwell94},
grammar modelling and induction \cite
{Atwell88b,Lari90,CarrollCharniak92,Atwell92,Bril92,Atwell93,JostAtwell94},
grammatical error detection \cite{Atwell88a,Atwell90},
probabilistic parsing
\cite {Sampsonetal89,SoODo91,Mage91,SoAt92,Atwell91,Bris92,Blac93}.
Particularly relevant to AMALGAM is the
recent research interest in Machine Translation using
statistical learning techniques for
mapping-extraction from parallel corpora
\cite{Brownetal90,Brownetal92,Chenetal91,WuXuanyin94}.

\section{Obtaining Resources}

As a development and testing resource, we are using the text of
the Lancaster-IBM
Spoken English Corpus (SEC) \cite{TaylorKnowles88}.
The SEC is a collection of recordings of radio
broadcasts with accompanying annotated transcriptions, collected by Lancaster
University and IBM UK as a general
research resource. The SEC is available from
the International Computer Archive of Modern English (ICAME) based at the
Norwegian Computing Centre for the Humanities (in Bergen, Norway). The corpus
exists in several forms and annotations: the digitised acoustic waveform; the
graphemic transcription annotated with prosodic markings; and a part-of-speech
analysis that was annotated semi-automatically with the aid of
CLAWS \cite{Atwell83,LeechGarsideAtwell83} as used for the LOB corpus.
Skeletal parsing has been added to
create the SEC Treebank, and this forms a subset of the Lancaster-IBM
Treebank. Gerry Knowles (Lancaster) and Peter Roach (Reading, formerly of
Leeds)
collaborated in an ESRC-funded project, MARSEC, to set up a
time-aligned database of recorded speech, accompanied by phonetic and
graphemic transcriptions \cite{Knowles93}.
Our proposal will produce, as a side-effect, several
alternative tagged and parsed versions of the SEC which will be made available
to the SEC database project collaborators. It will also be able to act as a
test-bed for the comparison and evaluation of parsing schemes.

Obtaining resources proved to be a stumbling block.  Whilst most  of the
people in charge of corpus annotation and
distribution are helpful they are also usually
very busy!  Sometimes there are reservations about distribution of
resources.  For example, the corpus could have copyright
restrictions or could be collected for dictionary compilation.
However, we have obtained the following corpora in tagged or parsed form
along with manuals defining the syntactic annotation schemes:
Brown \cite{FrancisKucera79},
LOB \cite{Atwell82,AtwellLeechGarside84,JohanssonAtwellGarsideLeech86},
London-Lund \cite{Svartvik90},
Polytechnic of Wales \cite{Souter89,FawcettPerkins80} and will apply for the
British National Corpus as soon as it becomes available.  We also have the
software used for annotating the
University of Pennsylvania corpus
\cite{BrillMarcus92,MarcusSantorini92}
and the International Corpus of English \cite{Greenbaum93,Barkema93}.

The following table summarises the resources we have for the six main
corpora we have included in the project so far.  The first column reveals
if we have the corpus itself: we have all but the International Corpus of
English.  The next column indicates if we have the software that was used
in the automated part of annotating of the corpus.  The next column shows for
which corpora we have documentation giving formal descriptions of the
annotation guidelines.  The last column marks the London-Lund and
Brown corpus with a `1' to indicate that we have a small sample of corpus
annotated using both these schemes. The `2' marker in this column indicates
the Parallel Annotated Corpus that we are building at the moment by adding
the International Ccorpus English (GB) annotation to the Spoken English Corpus.


\vspace{ 15pt }
\begin{tabular}{|l||c|c|c|c|} \hline
\multicolumn{5}{|c|}
{\bf Table 1: Summary of Resources} \\
\hline \hline
 & \multicolumn{4}{|c|}{\bf Do we have:} \\
\cline{2-5}
\multicolumn{1}{|c||}{\bf Corpora} &
{\bf corp} & {\bf soft} & {\bf doc} & {\bf PAC} \\ \hline \hline
Brown       & $\bullet$ &           &           & $\bullet_{1}$ \\
ICE         &           & $\bullet$ & $\bullet$ & $\bullet_{2}$ \\
LOB         & $\bullet$ & $\bullet$ & $\bullet$ &               \\
London-Lund & $\bullet$ &           &           & $\bullet_{1}$ \\
POW         & $\bullet$ &           &           &               \\
SEC         & $\bullet$ & $\bullet$ & $\bullet$ & $\bullet_{2}$ \\
\hline
\end{tabular}
\vspace{ 15pt }

\section{Deriving Tagset Mappings}

When we began the AMALGAM project we anticipated that the following process
would be the normal way that an annotation scheme was included in our
`mapping suite':

\begin{enumerate}
  \item Develop the most accurate mapping between the new scheme and
        one of the schemes already in the mapping suite.  Only one pair
        need to be mapped explicitly as the other  mappings can
        be generated from intermediaries
        via an `interlingua' approach \cite{AtwellHughesSouter94b}.
  \item Annotate the Spoken English Corpus using the mapping.
  \item Correct the mapped annotation, preferably using advice from the
        people responsible for the annotation scheme.
\end{enumerate}

The uneven spread of resources means that alternative mapping strategies must
be adopted when including each annotation scheme (see table 1). As we have the
software used to tag and parse the International Corpus of English we can
incorporate that into the mapping.  Good formal descriptions of the annotation
scheme (such as for LOB) can be used to craft some rules by hand.  Where the
documentation is sparse rules can be extracted from the corpus itself.

We require a method to evaluate the alternative mapping strategies: A simple
evaluation can be accomplished by tagging the untagged SEC using one
annotation scheme (the {\em evaluation} scheme) by the tried and tested method
of automatic annotation followed by hand correction.  To test a mapping
strategy
one would apply the mapping from the evaluation scheme tags to produce those
of the SEC.  The success of the mapping would be determined by measuring the
difference between this annotation and the original SEC (CLAWS tagged)
annotation produced by Lancaster.

The Parallel Annotated Corpus (PAC) created when a (non-CLAWS) evaluation
scheme is used to tag the Spoken English Corpus in this way itself provides
further possibilities for developing mapping strategies.  The PAC may
intrinsically encode mapping information that would not be uncovered from
other mapping strategies.  Extracting a mapping from a PAC is computationally
trivial; the difficulty is annotating an existing corpus with a new scheme.
However, PACs already exist for pairs of annotation scheme and this provides
an easy way to extract mapping information.  This is particularly true when
the annotation scheme of one corpus is replaced by another.  Initially this
would be done using the automatic annotator of the new scheme followed by
hand-correction by linguistic experts.  However, the addition of the new
scheme to part of the corpus creates a PAC from which a mapping can be
derived. The mapping could be used to update the performance of the automatic
annotator. A process of refinement of the automatic annotator by feedback
derived from the mapping would be established.

This paper focuses on deriving
tagset mappings from PACs as we are currently in
the phase of our project where we are concentrating on parts-of-speech
annotation.  However, we anticipate that the method will be even more useful
when dealing with mapping between parse trees.

\section{Extraction of Correspondences from Parallel Annotated Corpora}

Although a few PACs already exist
only a few tagset pairings are covered.
Often a corpus is annotated with a scheme that the designers
feel can be improved so they annotate the same texts with the updated
scheme.  This automatically results in a PAC being formed.
An example PAC comprises a few
sections of the Brown corpus that
were annotated by additional London-Lund markup \cite{Eeg-Olofsson91}.
A further example
is the Nijmegen Corpus which was originally annotated with CCPP annotation
\cite{Keulen86} but later replaced with the scheme used to
annotate the British component of the International Corpus of English
\cite{Greenbaum93}.
Although the Nijmegen TOSCA
team now view the CCPP scheme as largely obsolete it is still
a useful resource for mapping extraction as the PAC is 130,000 words in
length.  This provides a large sample from which to evaluate alternative
mapping strategies.

To use the method of deriving
mappings from PACs it is inevitable that some
traditional tagging is required to build the parallel corpus.  As an
example of the process of extracting correspondences from PACs
we shall use the example of the SEC-ICE corpus.  As a PAC does not exist for
this pair of tagsets we had to build our own.  As we aimed to produce the
multitagged corpus out of the texts of the Spoken English Corpus it
made sense to annotate the Spoken English Corpus with ICE tags.

We employed an experienced annotator of corpora, Tim Willis,
to learn the ICE annotation scheme and apply
it to the Spoken English Corpus by editing the automatic output of the Nijmegen
parser which was designed to annotate ICE-GB material.
For the moment
we are concentrating on deriving mappings between tagged annotation but it
was felt more cost effective to parse and tag the Spoken English Corpus now
as our project will eventually include parse mappings.

The output from the Nijmegen parser \cite{Barkema93}
needs to be aligned with the markup
in the Spoken English Corpus.  Problems are caused by the taggers segmenting
text by different methods.  Some taggers convert words not normally
capitalised into lowercase, but not all do.  This causes problems trying
to match the words again once annotation has taken place.
The Spoken English Corpus has sentence
boundaries after full stops, exclamation marks and question marks whereas
the Nijmegen parser additionally delimits text separated by colons and
semicolons.
The Nijmegen parser and The Spoken English Corpus tagging scheme deal with
enclitics in a similar manner; a word like {\em who's} being split into the
separate items {\em who} and {\em 's}.  Other schemes may leave such words
as they are.  To be aligned with the Spoken English Corpus would require
the word and its corresponding tag to to be split.  On the other hand,
a proper noun such as {\em New York} may be assigned a single tag and treated
as a single item rather than having the two words treated individually as in
the Spoken English Corpus.  The Nijmegen
parser does this when producing parsed
output but not when producing tagged output.  Some parsers alter the
text they annotate; again making the alignment process more difficult.  A
common practice is the removal of capital letters from words that would
not normally have them were they not starting a sentence.  Worse, the item
may be transformed altogether.  A semicolon found in the input to the
Nijmegen parser is transformed into the string {\em \&semi;} as the semicolon
on its own would be mistaken for an SGML marker \cite{Burnard91}.  Such issues
make alignment a non-trivial task.

\begin{verbatim}

  (             (
  In            IN
  Perspective   NP
  )             )
  (             (
  Rosemary      NP
  Hill          NP
  (             )
  -----         ---
  good          JJ     FRM:1/2
  morning       NN     FRM:2/2
  .             .      PUNC(per)
  -----         ---    -----
  more          AP     PRON(quant)
  news          NN     N(com,sing)
  about         IN     PREP(ge)
  the           ATI    ART(def)
  Reverend      NPT    N(prop,sing):1/4
  Sun           NP     N(prop,sing):2/4
  Myung         NP     N(prop,sing):3/4
  Moon          NP     N(prop,sing):4/4
  ,             ,      *PUNC(com)
  founder       NN     N(com,sing)
  of            IN     PREP(ge)
  the           ATI    ART(def)
  Unification   NNP    N(prop,sing):1/2
  church        NN     N(prop,sing):2/2
  ,             ,      *PUNC(com)
  who           WP     PRON(rel)
  's            BEZ    V(cop,pres,encl)
  currently     RB     ADV(ge)
  in            IN     PREP(ge)
  jail          NN     N(com,sing)
  for           IN     PREP(ge)
  tax           NN     N(com,sing):1/2
  evasion       NN     N(com,sing):2/2

   Figure 1: Alignment of SEC and ICE

\end{verbatim}

To align texts annotated by two schemes we used a method we term
{\em island driven alignment}.  The `islands' are the singletons found to be
present in the output of both schemes.  The position of these items can
easily be aligned.
The words next to the islands can be examined in turn.  Often
they will match and so can be aligned immediately, but occasionally the
next pair of items will not match.  Attempting to split enclitics, recombine
split compounds or altering initial letter case may match some pairs but
others such as the semicolon problem mentioned earlier will require pattern
matching of the surrounding text.
Occasionally an item in one of the annotations
will match with no item in the other; the extra end of sentence markers in ICE
texts being a good example.  When this happens it can only be discovered
after aligning the items on either side of it with neighbouring items in
the other annotated output.  The first few lines of the Spoken English Corpus
when aligned with the ICE tags of the same text are shown figure 1, above.
The first
two columns are the words and CLAWS tags from the tagged SEC and the remaining
column contains the corresponding ICE tags.

The Spoken English Corpus contains the short header:
{\em (In Perspective)(Rosemary Hill)}.  The process by which ICE was annotated
excluded headers such as this (they will be tagged by hand).  As the header
is not included in the ICE annotation of the text there is nothing to align it
to.

Each pairing of tags can now be counted and a list of correspondences
made for each individual tag to show the probabilities of each pair.  For
instance the London-Lund/Brown PAC produced the list of London-Lund
correspondences for the interrogative wh-determiner tag, {\em WDT}, in Brown
shown in figure 2.

\begin{verbatim}

                  B2deg    2.13%
                  BHitr   25.53%
      WDT x --->  BRwha    4.26%
                  GAwhi   53.19%
                  GCwha   14.89%

    Figure 2: Correspondences for WDT

\end{verbatim}

The Brown tag {\em WDT} pairs with the London-Lund tag {\em GAwhi},
relative pronoun: {\em which}, just over
half the time in the PAC.  The easiest way to convert these correspondences
into a mapping is to map the tag in one scheme always onto the most common
pairing found in the PAC.  Many tags will have a 1:1 mapping or will pair
with one particular tag in the other scheme almost all the time.  However,
the above example correspondence list illustrates where mapping the most
common pairing will work badly.  We are currently investigating methods
of incorporating the lexicon (which could be extracted from the corpus
samples we have, or from the PACs we have built ourselves) or using the
contextual information supplied by the neighbouring words and tags.
We also hope to explore methods developed by Brill in which texts were
first tagged
by always selecting the most common tag for a word, and then
the tag selection refined with a set of automatically extracted
rewrite rules, or {\em patches} \cite{Brill91}.

\section{Lessons for the EAGLES Initiative}

Until recently, very little effort has been expended on the development of
standards in tagging and parsing natural language corpora. Individual tagging
and parsing schemes have been invented
more or less independently, and differ not
only in the linguistic description, but also in the formalism used to label
words or represent tree structures. \cite{Souter93} surveys some of the
substantive differences between such formalisms for contemporary
parsed corpora of English, and illustrates how standards are needed
to facilitate the reusability of corpus resources (through enterprises such as
the Text Encoding Initiative),
and to improve
the general applicability of corpus-processing software, such as the Nijmegen
Linguistic DataBase \cite{vanHalteren90}.

As many participants at the workshop will know, EAGLES is a European
initiative to devise a set of common standards for Natural Language
Processing technology across the range of European Union working languages.
Of particular relevance to our research are the standards proposals for
morphosyntactic wordclasses; a lengthy draft proposal (over 200 pages)
has recently been made available to ELSNET nodes and a number of other
centres of expertise for comment. The proposals aim to standardise a set of
wordclasses to be applied to Danish, Dutch, English, French, German, Greek,
Italian, Portuguese, and Spanish; once (or if) agreed, the standards may
later be extended to cover other languages (e.g. Swedish, Finnish,
Norwegian, Gaelic, Welsh, Basque, \ldots)  Even among the current EU main
languages, there is considerable diversity in morphosyntax, so the EAGLES
group are to be congratulated for achieving a compromise which on the face
of it is largely uncontentious.  EAGLES recommends several levels of
refinement or delicacy in wordclasses, so that specific applications
and/or language models are free to select an appropriate level of
tagset granularity.  For example, {\em NOUN} is a broad (level 1) category,
a general class which all language models must recognise; within this,
there is a level 2 subdivision into proper nouns and common nouns,
which will apply to many but not all applications etc.  Many other
possible wordclass distinctions are captured by features, e.g. number,
gender; some of these do not apply to certain languages (eg gender of
English nouns).

Unfortunately, the divisions between word classes and subclasses are
made in terms of examples, and appeals to linguistic intuition.  This is
reasonable and normal practice in lexicography and language teaching;
but for computational implementation definitions and boundaries need to be
more clearly specified.  Otherwise, there is a danger that NLP systems
will adopt wordclass-demarcations on grounds of computational tractability,
which may not agree with the linguistically correct/intuitive definition.
Worse still, although linguists agree on the general "common-sense"
definitions of categories like proper noun, common noun etc, our
analysis of competing tagsets for English corpora shows that these
categories are in fact `fuzzy', and different corpus tagging projects
have adopted subtly but significantly different definitions, probably
unaware that their analyses are incompatible with those of other linguists.
The EAGLES recommendations include a call to corpus tagging projects to
provide their manuals or tagset-definitions along with the final tagged
corpus, but we have found that, to date, tagging project teams have deemed
these `case-law' handbooks as `training in progress statements'
not worth publishing - with the notable exception
of \cite{JohanssonAtwellGarsideLeech86}.

Our earlier example of parallel CLAWS/ICE tagging of the Spoken English
Corpus illustrates the fuzziness in the distinction between proper
noun and common noun.  In general, a singular proper noun is NP in LOB
and CLAWS, but N(prop,sing) in ICE.  However, notice
that {\em Perspective}, the second
word in the corpus, is tagged NP.  This may have been because the word begins
with a capital, and the tagging system uses this as a deciding criterion
(however, note that the previous
word, {\em In}, escapes this default NP tagging because English text
requires the first word of every sentence to start with a capital, so the
tagging system by default converts this to lower case and tags according to
dictionary-lookup).
To a linguist, this analysis of {\em Perspective} may intuitively
be an `error; however there are no definitions within the EAGLES guidelines
which rule out such counter-intuitive computationally-motivated criteria.

A second example of disagreement over the proper and common noun boundary
is the analysis of {\em Reverend Sun Myung Moon} - in ICE this
is tagged as a proper-noun sequence (or rather, a compound proper-noun
single lexical item), but in
LOB/CLAWS, one fuzzy boundary between common and proper nouns is
recognised - the area of titular nouns tagged NPT (for example, {\em Reverend}
can start with upper or lower case in much the same context, so NPT avoids
conflicting taggings depending on the case of the initial letter).
Further examples abound in the parallel corpus; generally the problem
arises from differences in the handling of upper-case initial letter.

Our conclusion for the EAGLES Initiative is that the morphosyntactic
category proposals must be followed up with detailed definitions,
preferably including computable criteria. In the specific example of
nouns, there must be clear standards on handling of word-initial case.
(This is relevant not only to English).  Otherwise the `standards'
will be interpreted differently (and incompatibly) in different tagged
corpora.  We had hoped that the EAGLES tagset might constitute
an `interlingua' for translating between existing tagsets. However,
we have already had to conclude
that our task of automatic tagset-mapping extraction
can never achieve perfect accuracy, as both source and target training
data are noisy; using a fuzzy-edged tagset as an interlingua could only
worsen matters.

\bibliographystyle{acl}

\begin{thebibliography}{}

\bibitem[Atwell 82]{Atwell82}
Eric Atwell.
\newblock 1982.
\newblock {\em LOB corpus tagging project: Manual post-edit handbook}.
\newblock Departments of Computer Studies and Linguistics,
Lancaster University.

\bibitem[Atwell 83]{Atwell83}
Eric Atwell.
\newblock 1983.
\newblock {\em Constituent likelihood grammar}.
\newblock {\em ICAME Journal}, {\bf 7}.  34--67.

\bibitem[Atwell 87a]{Atwell87a}
Eric Atwell.
\newblock 1987a.
\newblock Constituent likelihood grammar.
\newblock In Roger Garside, Geoffrey Sampson and Geoffrey Leech (eds.),
{\em The computational analysis of English: A corpus-based approach}.  57--65.

\bibitem[Atwell 87b]{Atwell87b}
Eric Atwell.
\newblock 1987b.
\newblock A parsing expert system which learns from corpus analysis.
\newblock In Willem Meijs (ed.), {\em Corpus Linguistics and Beyond:
  Proceedings of the ICAME 7$^{th}$ International Conference}.
\newblock Amsterdam: Rodopi.  227--235.

\bibitem[Atwell 88a]{Atwell88a}
Eric Atwell.
\newblock 1988a.
\newblock Grammatical analysis of English by statistical pattern recognition.
\newblock In Josef Kittler (ed.), {\em Pattern recognition: Proceedings of
the 4$^{th}$ International Conference, Cambridge}.
\newblock Berlin: Springer-Verlag.   626--635.

\bibitem[Atwell 88b]{Atwell88b}
Eric Atwell.
\newblock 1988b.
\newblock Transforming a parsed corpus into a corpus parser.
\newblock In Merja Kyto, Ossi Ihalainen, and Matti Risanen (eds.),
{\em Corpus Linguistics, hard and soft: Proceedings of the ICAME 8th
International Conference}.
\newblock Amsterdam: Rodopi.  61--70.

\bibitem[Atwell 90]{Atwell90}
Eric Atwell.
\newblock 1990.
\newblock Measuring grammaticality of machine-readable text.
\newblock In Werner Bahner, Joachim Schildt, and Dieter Viehweger (eds.),
{\em Proceedings of the Fourteenth International Congress of Linguists},
  {\bf III}.
\newblock Berlin: Akademie-Verlag.  2275--2277.

\bibitem[Atwell 92]{Atwell92}
Eric Atwell.
\newblock 1992.
\newblock Overview of grammar acquisition research.
\newblock In Henry Thompson (ed.), {\em Workshop on sublanguage grammar and
lexicon acquisition for speech and language: Proceedings}.  65--70.

\bibitem[Atwell 93]{Atwell93}
Eric Atwell.
\newblock 1993.
\newblock Corpus-based statistical modelling of English grammar.
\newblock In Souter and Atwell (eds.), {\em Corpus-based computational
  linguistics.}
\newblock Amsterdam: Rodopi.  195--215.

\bibitem[Atwell et al 84]{AtwellLeechGarside84}
Eric Atwell, Geoffrey Leech, and Roger Garside.
\newblock 1984.
\newblock Analysis of the LOB corpus: Progress and prospects.
\newblock In Jan Aarts and Willem Meijs (eds.), {\em Corpus linguistics:
Proceedings of the ICAME 4$^{th}$ International Conference on the Use
of Computer Corpora in English Language Research}.  Amsterdam: Rodopi.

\bibitem[Atwell and Drakos 87]{AtwellDrakos87}
Eric Atwell and Nikos Drakos.
\newblock 1987.
\newblock Pattern recognition applied to the acquisition of a grammatical
classification system from unrestricted English text.
\newblock In Bente Maegaard (ed.), {\em Proceedings of the Third Conference
of the European Chapter of the Association for Computational Linguistics}.
\newblock New Jersey, USA.  56--63

\bibitem[Atwell et al 91]{Atwell91}
Eric Atwell, Clive Souter and, Tim O'Donoghue.
\newblock 1991.
\newblock {\em Training {P}arsers with {P}arsed {C}orpora:  Report 91.2.}
\newblock School of Computer Studies, University of Leeds, UK.

\bibitem[Atwell et al 94a]{AtwellHughesSouter94a}
Eric Atwell, John Hughes, and Clive Souter.
\newblock 1994a.
\newblock A unified multicorpus for training syntactic constraint models.
\newblock In Lindsay Evett and Tony Rose (eds.), {\em Proceedings of
AISB Workshop on Computational Linguistics for Speech and
Handwriting Recognition}.
\newblock Leeds University, UK.

\bibitem[Atwell et al 94b]{AtwellHughesSouter94b}
Eric Atwell, John Hughes, and Clive Souter.
\newblock 1994b.
\newblock AMALGAM: Automatic mapping among lexico-grammatical annotation
models.
\newblock In Judith Klavans and Philip Resnik (eds.),
{\em Proceedings of the balancing act - combining
symbolic and statistical approaches to language},
\newblock Workshop in Conjunction with the 32nd Annual Meeting of the
Association for Computational Linguistics.
\newblock New Mexico State University, Las Cruces,
New Mexico, USA.

\bibitem[Barkema 93]{Barkema93}
Henk Barkema.
\newblock 1993.
\newblock {\em The TOSCA Analysis Environment for ICE.}
\newblock TOSCA, University of Nijmegen, The Netherlands.

\bibitem[Black et al 93]{Blac93}
Ezra Black, Roger Garside, and Geoffrey Leech (eds.).
\newblock 1991.
\newblock {\em Statistically driven computer grammars of {E}nglish: The
  {IBM}-{L}ancaster approach}.
\newblock Rodopi.

\bibitem[Brill 91]{Brill91}
Eric Brill.
\newblock 1991.
\newblock {\em A simple rule-based part of speech tagger.}
\newblock Technical report: Department of Computer Science, University of
Pennsylvania.

\bibitem[Brill et al 92]{Bril92}
Eric Brill, David Magerman, Mitchell Marcus, and Beatrice Santorini.
\newblock 1992.
\newblock Deducing linguistic structure from the statistics of large corpora.
\newblock In {\em Proceedings of the AAAI-92 Workshop on
Statistically-Based NLP {T}echniques}.
\newblock San Jose, California, USA.

\bibitem[Brill and Marcus 92]{BrillMarcus92}
Eric Brill and Mitchel Marcus.
\newblock 1992.
\newblock Tagging an unfamiliar text with minimal human supervision.
\newblock In Robert Goldman (ed.),
{\em Working Notes of the AAAI Fall Symposium on Probabilistic Approaches to
Natural Language}, AAAI Press.

\bibitem[Briscoe and Waegner 92]{Bris92}
Ted Briscoe and Nick Waegner.
\newblock 1992.
\newblock Robust stochastic parsing using the {I}nside-{O}utside {A}lgorithm.
\newblock In {\em Proceedings of the AAAI-92 Workshop on Statistically-Based
NLP Techniques}.
\newblock San Jose, California, USA.

\bibitem[Brown et al 90]{Brownetal90}
Peter Brown, John Cocke, Stephen DellaPietra, Vincent DellaPietra,
Frederik Jelinek, John Laffety, Robert Mercer, Paul Roosin.
\newblock 1990.
\newblock A statistical approach to machine translation.
\newblock {\em Computational Linguistics}, {\bf 16}.  29--85.

\bibitem[Brown et al 92]{Brownetal92}
Peter Brown, Stephen DellaPietra, Vincent DellaPietra, John Laffety,
Robert Mercer.
\newblock 1992.
\newblock Analysis, statistical transfer, and synthesis in machine
translation.
\newblock in {\em Fourth International Conference on Theoretical and
Methodological Issues in Machine Translation}.
\newblock Montreal.  83--100.

\bibitem[Burnard 91]{Burnard91}
Lou Burnard.
\newblock 1991.
\newblock What is the TEI?
\newblock In D. Greenstein (ed.), {\em Modelling historical data.}
\newblock Goettingen: St. Katharinen.

\bibitem[Carroll and Charniak 92]{CarrollCharniak92}
Glenn Carroll and Eugene Charniak.
\newblock 1992.
\newblock Two experiments on learning probabilistic dependency grammars from
corpora.
\newblock In {\em Proceedings of the AAAI-92 Workshop on Statistically-Based
NLP Techniques}.
\newblock San Jose, California, USA.  1--13.

\bibitem[Chen et al 91]{Chenetal91}
S.-C. Chen, J.-S. Chang, J.-N. Wang, and K.-Y. Su.
\newblock 1991.
\newblock ArchTran: A corpus-based statistics-oriented English-Chinese machine
translation system.
\newblock In {\em Proceedings of Machine Translation Summit III}.
\newblock Washington, D.C.  33--40.


\bibitem[Demetriou and Atwell 93]{Demetriou93}
George Demetriou and Eric Atwell.
\newblock 1993.
\newblock Machine-learnable, non-compositional semantics for domain
independent speech or text recognition.
\newblock In {\em Proceedings of 2$^{nd}$ Hellenic-European Conference
on Mathematics and Informatics (HERMIS)}.
\newblock Athens University of Economics and Business,  Greece.

\bibitem[Eeg-Olofsson 91]{Eeg-Olofsson91}
Mats Eeg-Olofsson.
\newblock 1991.
\newblock {\em Word-class tagging - some computational tools.}
\newblock G\"{o}teborgs Universitet Institutionen f\"{o}r
Spr\aa kvetenskaplig Databehandling.

\bibitem[Fawcett and Perkins 80]{FawcettPerkins80}
Robin Fawcett and Michael Perkins.
\newblock 1980.
\newblock {\em Child language transcripts 6-12.
(With a preface, in 4 volumes).}
\newblock Department of Behavioural and Communication Studies,
Polytechnic of Wales.

\bibitem[Francis and Ku\v{c}era 79]{FrancisKucera79}
W.N. Francis and H. Ku\v{c}era.
\newblock 1979.
\newblock {\em Manual of information to accompany a standard corpus of
present-day edited American English, for use with digital computers
(corrected and revised edition).}
\newblock Department of Linguistics, Brown University,
Providence, Rhode Island.

\bibitem[Gale et al 92]{Galeetal92}
William Gale, Kennethe Church, and David Yarowsky.
\newblock 1992.
\newblock Using bilingual materials to develop word sense disambiguation
methods.
\newblock In {\em Fourth International Conference on Theoretical and
Methodological Issues in Machine Translation}.
\newblock Montreal.  101--112.

\bibitem[Greenbaum 93]{Greenbaum93}
Sidney Greenbaum.
\newblock 1993.
\newblock The tagset for the International Corpus of English.
\newblock In Clive Souter and Eric Atwell (eds.),
{\em Corpus-based Computational Linguistics}.
Amsterdam: Rodopi.

\bibitem[van Halteren and van den Heuvel 90]{vanHalteren90}
Hans van Halteren and Theo van den Heuvel.
\newblock 1990.
\newblock {\em Linguistic Exploitation of Syntactic Databases}.
\newblock Amsterdam: Rodopi.

\bibitem[Hughes and Atwell 93]{Hughes93}
John Hughes and Eric Atwell.
\newblock 1993.
\newblock Acquiring and evaluating a classification of words.
\newblock In Simon Lucas (ed.), {\em {IEE} Grammatical Inference Colloquium}.
\newblock University of Essex, Colchester, UK.

\bibitem[Hughes and Atwell 94]{HughesAtwell94}
John Hughes and Eric Atwell.
\newblock 1994
\newblock The automated evaluation of inferred word classifications.
\newblock  In Tony Cohn (ed.), {\em The 11$^{th}$ European Conference on
Artificial Intelligence}.
\newblock RAI Congress Centre, Amsterdam, The Netherlands.

\bibitem[Johansson et al 86] {JohanssonAtwellGarsideLeech86}
Stig Johansson, Eric Atwell,  Roger Garside, and  Geoffrey Leech.
\newblock 1986.
\newblock {\em The tagged LOB corpus: Users' manual.}
\newblock The Norwegian Centre for the Humanities, Bergen.

\bibitem[Jost and Atwell 94]{JostAtwell94}
Uwe Jost and Eric Atwell.
\newblock 1994.
\newblock Capturing long-distance syntactic constraints in a bigram model.
\newblock In  Lindsay Evett and Tony Rose (eds.), {\em Proceedings of AISB
Workshop on Computational Linguistics for Speech and Handwriting Recognition}.
\newblock Leeds University, UK.

\bibitem[Keulen 86]{Keulen86}
Fran\c{c}oise Keulen.
\newblock 1986.
\newblock The Dutch Computer Corpus Pilot Project.
\newblock In Jan Aarts and Willem Meijs (eds.),
{\em Corpus Linguistics II},
Amsterdam: Rodopi.  127--163.

\bibitem[Knowles 93]{Knowles93}
Gerry Knowles.
\newblock 1993.
\newblock From text to waveform: Converting the Lancaster/IBM Spoken
English Corpus into a speech database.
\newblock In Clive Souter and Eric Atwell (eds.),
{\em Corpus-Based Computational Linguistics}.
Amsterdam: Rodopi.  47--58.

\bibitem[Lari and Young 90]{Lari90}
K.~Lari and S.~J. Young.
\newblock 1990.
\newblock The estimation of stochastic context-free grammars using the
  {I}nside-{O}utside {A}lgorithm.
\newblock {\em Computer {S}peech and {L}anguage}, {\bf 4}.  35--56.

\bibitem[Leech et al 83]{LeechGarsideAtwell83}
Geoffrey Leech, Roger Garside, and Eric Atwell.
\newblock 1983.
\newblock The automatic grammatical tagging of the LOB Corpus.
\newblock {\em ICAME Journal}, 7:13--33.

\bibitem[Leech and Garside 91]{LeechGarside91}
Geoffrey Leech and Roger Garside.
\newblock 1991.
\newblock Running a grammar factory: The production of syntactically-annotated
corpora or `treebanks'.
\newblock In Stig Johannsson and Anna-Brita Strenstr\"{o}m (eds.),
{\em English Computer Corpora}.
Berlin: Mouton de Gruyter.  15--32.

\bibitem[Magerman and Marcus 91]{Mage91}
D.~Magerman and M.~Marcus.
\newblock 1991.
\newblock Pearl: A probabilistic chart parser.
\newblock In {\em Proceedings of the $2^{nd}$ International Workshop on
Parsing Technologies}.
\newblock Cancun, Mexico. 193--199.

\bibitem[Marcus and Santorini 92]{MarcusSantorini92}
Mitchel Marcus and B. Santorini.
\newblock 1992.
\newblock Building very large natural language corpora:  The Penn treebank.
\newblock In N. Ostler (ed.),
{\em Proceedings of the 1992 Pisa Symposium on European Textual Corpora.}

\bibitem[Sampson et al 89]{Sampsonetal89}
Geoffrey Sampson, Robin Haigh, and Eric Atwell.
\newblock 1989.
\newblock Natural language analysis by stochastic optimisation: A progress
  report on project April.
\newblock {\em Journal of Experimental and Theoretical Artificial
  Intelligence}, {\bf 1}.  271--287.

\bibitem[Souter 89]{Souter89}
Clive Souter.
\newblock 1989.
\newblock {\em A short handbook to the Polytechnic of Wales corpus.}
\newblock Bergen: Norwegian Computing Centre for the Humanities,
Bergen University.

\bibitem[Souter 93]{Souter93}
Clive Souter.
\newblock 1993.
\newblock Towards a standard format for parsed corpora.
\newblock In J. Aarts, P. de Haan and N. Oostdijk (eds.),
{\em English Language Corpora: Design, Analysis and Exploitation.}
Amsterdam: Rodopi.  197--214.

\bibitem[Souter and O'Donoghue 91]{SoODo91}
Clive Souter and Tim O'Donoghue.
\newblock 1991.
\newblock Probabilistic parsing in the communal project.
\newblock In Stig Johansson and Anna-Brita Stenstrom (eds.), {\em English
Computer Corpora, Selected Papers and Research Guide}.
\newblock Berlin: Mouton de Gruyter.  33--48.

\bibitem[Souter and Atwell 92]{SoAt92}
Clive Souter and Eric Atwell.
\newblock 1992.
\newblock A richly annotated corpus for probabilistic parsing.
\newblock In {\em In Proceeding of the AAAI-92 Workshop on
Statistically-Based NLP Techniques}.
\newblock San Jose, California, USA.

\bibitem[Svartvik 90]{Svartvik90}
Jan Svartvik (ed.).
\newblock 1990.
\newblock {\em The London-Lund corpus of spoken English:
Description and Research.}
\newblock Lund University Press, Lund, Sweden.

\bibitem[Taylor and Knowles 88]{TaylorKnowles88}
L.J. Taylor and G. Knowles.
\newblock 1988.
\newblock {\em Manual of information to accompany the SEC corpus.}
\newblock Technical report, Unit for Computer Research on the English Language,
University of Lancaster, UK.

\bibitem[Wu and Xia 94]{WuXuanyin94}
Dekai Wu and Xuanyin Xia.
\newblock 1994.
\newblock Learning an English-Chinese lexicon from a parallel corpus.
\newblock In {\em AMTA-94, Association for Machine Translation in the
Americas}.
\newblock Columbia, Maryland, USA.

\end{thebibliography}

\end{document}